\documentclass{article}
\usepackage{spconf,amsmath,graphicx}

\usepackage{algorithm}
\usepackage{algorithmic}
\usepackage{subcaption}
\usepackage{amssymb}
\usepackage{adjustbox}
\usepackage{newfloat}
\usepackage{listings}
\usepackage{xspace}
\usepackage{booktabs}

\usepackage{hyperref}
\usepackage{colortbl}

\newcommand{\model}{\textit{LaCViT}\xspace}
\title{LaCViT: A Label-aware Contrastive Fine-tuning Framework for Vision Transformers}
\name{Zijun Long, Richard McCreadie, Gerardo Aragon Camarasa, Zaiqiao Meng}
\address{The University of Glasgow, Scotland, UK}
\begin{document}
%
\maketitle
\begin{abstract}
\looseness -1 Vision Transformers (ViTs) have emerged as popular models in computer vision, demonstrating state-of-the-art performance across various tasks. This success typically follows a two-stage strategy involving pre-training on large-scale datasets using self-supervised signals, such as masked random patches, followed by fine-tuning on task-specific labeled datasets with cross-entropy loss. However, this reliance on cross-entropy loss has been identified as a limiting factor in ViTs, affecting their generalization and transferability to downstream tasks. Addressing this critical challenge, we introduce a novel Label-aware Contrastive Training framework, \model{}, which significantly enhances the quality of embeddings in ViTs. \model{} not only addresses the limitations of cross-entropy loss but also facilitates more effective transfer learning across diverse image classification tasks. Our comprehensive experiments on eight standard image classification datasets reveal that \model{} statistically significantly enhances the performance of three evaluated ViTs by up-to 10.78\% under Top-1 Accuracy. 
\end{abstract}
\begin{keywords}
Contrastive Learning, Vision Transformers, Fine Tuning, Transfer Learning,
\end{keywords}

\section{Introduction}
\label{sec:intro}

Transformers have significantly advanced the field of computer vision, particularly in tasks such as image classification~\cite{RN52,RN83,RN82,longautocrisis,longmultimodal,long2023robollm,yi2023large}. These models typically follow a two-stage process: pre-training on auxiliary tasks and fine-tuning on specific tasks using cross-entropy loss. However, the reliance on cross-entropy often leads to poor generalization and vulnerability to label noise and adversarial attacks~\cite{DBLP:conf/icml/LiuWYY16, DBLP:journals/corr/abs-1906-07413, DBLP:journals/corr/abs-1901-08360,10095195,long2023hard,long2023elucidating}, which impede their efficacy in practical applications. Moreover, vision transformers exhibit a lack of learned inductive biases~\cite{RN83,long2023crisisvit}, an essential feature for handling unseen examples and enhancing transfer learning. 

\looseness -1 The inherent lack of inductive bias and the limitations of fine-tuning with cross-entropy compromise the transfer learning capabilities of vision transformers, particularly when the target domain has a small sample size~\cite{zhou2021convnets}. Although previous works have attempted to address these issues by integrating convolutional neural networks or modifying the transformer architecture~\cite{RN102,RN103,RN105}, these solutions often compromise the inherent advantages of transformers, like their training efficiency and scalability. Hence, it would be advantageous to have an alternative approach to improve the transfer effectiveness of vision transformers without relying on convolutional models or layers, while utilizing task labels in the fine-tuning stage.

In response, we propose the \model{}, a label-aware contrastive training framework designed specifically for vision transformers. \model{} leverages task labels in a contrastive learning context to fine-tune pre-trained models, thus significantly improving their transfer learning capabilities. This framework employs a two-stage, label-aware contrastive learning loss to refine sample embeddings, enabling better generalization to target tasks. Notably, \model{} is the first framework of its kind to enhance vision transformer transfer learning without relying on convolutional layers or extended training epochs. We believe that our work serves as an impetus for the research community to reconsider the fine-tuning mechanisms for Vision Transformers. The primary contributions of this work are as follows:

\begin{itemize}
    \item We introduce \model{}, a pioneering label-aware contrastive fine-tuning framework that substantially boosts the transfer learning capabilities of vision transformers.
    \item We demonstrate the wide applicability of \model{} by fine-tuning three vision transformer models using it.
    \item Extensive experimentation across eight image classification datasets shows that \model{} significantly outperforms baseline models, such as a 10.78\% increase in Top-1 Accuracy for the \model{}-trained MAE on the CUB-200-2011 dataset~\cite{RN82}.
    \item Additional analysis shows that \model{} effectively reshapes pretrained embeddings into a more discriminative space, enhancing performance on target tasks.
\end{itemize}

\begin{figure*}
\centering
\vspace{-4mm}
\includegraphics[height=6.5cm]{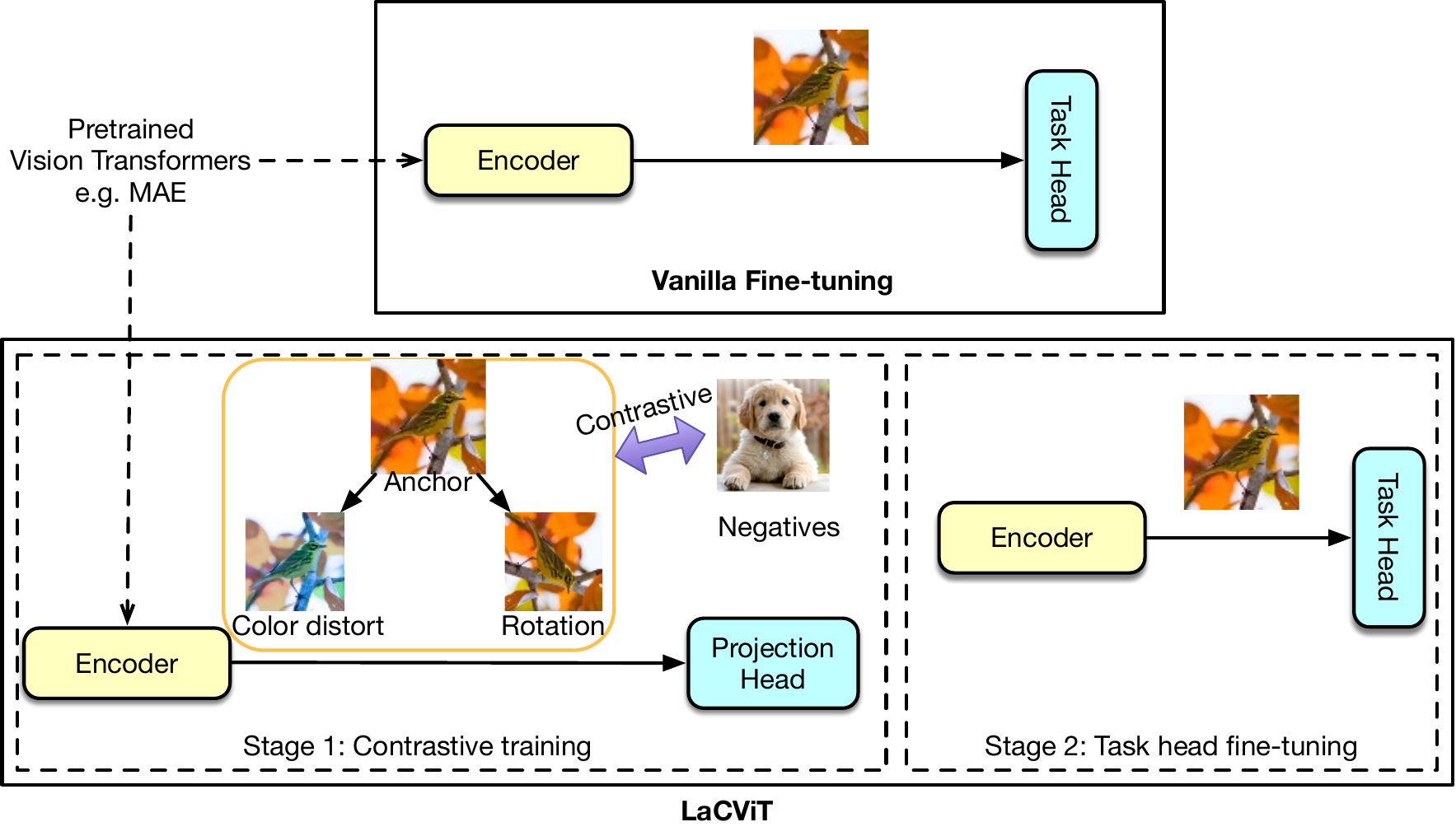}
\caption{\textbf{The overview of \model, which consists of two training stages: 1) label-aware contrastive training and 2) task head fine-tuning}, compared to the vanilla fine-tuning, which directly fine-tunes the task head. The first contrastive training stage trains the vision transformers based on the labels of the target tasks with a contrastive loss, aiming to improve the embedding quality, and in the second stage, \model{} is fine-tuned with a task-specific head.}
\label{fig:all_training_flow}
\vspace{-4mm}
\end{figure*}

\section{Related Work}
\vspace{-2mm}
\label{related_work}

\subsection{Vision Transformers}
Vision Transformers (ViTs) mark a significant advance when tackling image-based tasks, exemplified by the seminal ViT model~\cite{RN83}. Early works primarily focused on localized processing of image segments~\cite{RN93, RN94}. Subsequent developments, such as Masked Autoencoders (MAE), tackled the computational challenges of global attention mechanisms by employing high image masking strategies ~\cite{RN82}. Explorations in various pretraining methods, including SimMIM~\cite{xie2022simmim} and Data2vec~\cite{data2vec}, further advanced the field. Despite these advancements, a common challenge persists in the limited transferability of these models, particularly evident when employing cross-entropy loss for fine-tuning~\cite{RN83,zhou2021convnets}.
\vspace{-2mm}
\subsection{Contrastive Learning}
The development of contrastive learning trace back to early explorations by Becker~\cite{becker1992self}. This approach aims to differentiate similar items from dissimilar ones within an embedding space. Additionally, contrastive learning has shown remarkable efficacy in improving deep learning model performance across various domains~\cite{ long2023robollm,long2023multiwayadapater}, including sentence~\cite{10095914,10096142} and audio representation learning~\cite{10095373}, with its most notable impact observed in image recognition tasks, as exemplified by SimCLR~\cite{RN89} and other studies~\cite{ long2023hard,long2023elucidating,long2023crisisvit}. While integrating label information into contrastive learning has been explored, as in \cite{RN81}, these efforts have primarily remained confined to the pre-training phase and have not been extended to vision transformers.
\vspace{-2mm}
\subsection{Knowledge Gap and Our Contribution}
While both fields have advanced in parallel, the integration of label-aware contrastive learning within the fine-tuning stage of vision transformers remains unexplored. Our work addresses this gap by pioneering the application of contrastive learning during the fine-tuning phase of vision transformers, thereby enhancing their transferability.

\section{The Proposed Approach}
\label{sec:method}

\subsection{Motivation and Overview}
The prevalent approach for fine-tuning vision transformers employs cross-entropy loss, which suffers from poor generalization capabilities. Moreover, existing contrastive learning methods overlook the utility of label information in the fine-tuning phase. To address these limitations, we propose \model{}, a label-aware contrastive fine-tuning framework designed to enhance the transfer learning capabilities of vision transformers. As illustrated in Figure \ref{fig:all_training_flow}, our \model{} consists of two distinct stages: \emph{the label-aware contrastive training stage} and \emph{the task head fine-tuning stage}.

\begin{itemize}
\vspace{-1mm} 
    \item \textbf{Label-aware Contrastive Training Stage (Stage 1)}: In this stage, we initialize the model with pretrained weights and adapt these weights through a contrastive learning loss that incorporates the label information of the target task. This process comprises four main steps: data augmentation, patch encoding, nonlinear projection, and contrastive loss computation.
    \vspace{-1mm} 
    \item \textbf{Task Head Fine-tuning Stage (Stage 2)}: The second stage focuses on training the task-specific head, typically a simple linear layer for classification tasks. This stage utilizes a standard cross-entropy loss to fine-tune the whole framework, which is added atop the pretrained vision transformer.
    \vspace{-1mm} 
\end{itemize}

\subsection{Label-aware Contrastive Training Stage}
\label{sec:contrastive}

\noindent\textbf{Data Augmentation}: To augment each image in the mini-batch into two transformed views, we employ AutoAugment~\cite{DBLP:conf/cvpr/CubukZMVL19}, which has proven to be highly effective for contrastive learning.

\vspace{1mm} \noindent\textbf{Encoding}: Feature embeddings for each of the two augmented views of the image are generated using an encoder, such as ViT~\cite{RN83}, MAE~\cite{RN82}, or SimMIM~\cite{xie2022simmim}.

\vspace{1mm} \noindent\textbf{Nonlinear Projection Head}: To enhance the quality of the embeddings, we employ a nonlinear projection head, \( g(\boldsymbol{h}) \), upon the encoder to map the representation to the space where the contrastive loss is applied. Thus, to implement \( z_i = g(\boldsymbol{h_i}) = W^{(2)}\sigma(W^{(1)}\boldsymbol{h_i}) \), we use two dense layers, where \( \sigma \) is a ReLU function. The \( z = g(\boldsymbol{h}) \) is trained to be invariant to data transformation, which means \( g \) removes information that could be useful for the downstream task (e.g., color of objects). By using the nonlinear projection, more information can be maintained in \( h \). These embeddings are then grouped into distinct sets by the training class label of the source image.

\vspace{1mm} \noindent\textbf{Contrastive Loss Objective}: To obtain a more discriminative representation space for the target task, we train the pretrained encoder using a contrastive loss by leveraging label information (i.e., the label-aware contrastive loss). The label-aware contrastive loss enables stronger geographic clustering of samples belonging to the same class in the embedding space, while simultaneously pushing apart clusters of samples from different classes. The advantage of the label-aware contrastive loss is that we compute the contrastive loss based on true positive pairs per anchor in addition to true negative samples, compared to self-supervised contrastive learning that uses only augmented views. The contrastive loss is mathematically defined as follows:

\begin{equation}
\resizebox{0.43\textwidth}{!}{$
\vspace{-1.5mm} 
\mathcal{L}(\mathcal{D}^{*}) = \displaystyle\sum\limits_{z_{i}\in \mathcal{D}^{*}} \frac{-1}{\lvert \mathcal{D}^{*\textbf{+}}_{-z_i}\lvert} \displaystyle\sum\limits_{z_{p}\in \mathcal{D}^{*\textbf{+}}_{-z_i}} \log\frac{\exp(z_{i}\cdot z_{p}/\tau)}{\displaystyle\sum\limits_{z_{a}\in \mathcal{D}^{*}_{-z_i}} \exp(z_{i}\cdot z_{a}/\tau)},
$}
    \label{eq:contrastive_loss_fn}
\vspace{-0.5mm}    
\end{equation}

In Equation~\ref{eq:contrastive_loss_fn}, $\mathcal{D}^{*}$ represents the entire mini-batch composed of an embedding $z$ for each image view (or anchor) $i$. Therefore, $z_i\in\mathcal{D}^{*}$ is a set of embeddings within the mini-batch.  The superscripts $+$ and $-$, e.g. $\mathcal{D}^{*\textbf{+}}$, denote sets of embeddings consisting only of positive and negative examples, respectively, for the current anchor within the mini-batch. The term $\lvert \mathcal{D}^{*\textbf{+}}_{-z_i}\lvert$ represents the cardinality of the positive set for the current anchor, while the subscript $-z_i$ denotes that this set excludes the embedding $z_i$. The symbol $\cdot$ represents the dot product. $\tau$ is a temperature parameter, which controls the degree of loss applied when two images have the same class but the embeddings are different. A higher value pushes the model to more strongly separate the positive and negative examples. 

\vspace{1mm} \noindent \textbf{Comparison with Existing Methods}. Different from previous methods that integrate convolutional layers into the transformer architecture or extend the training epochs, our \model{} preserves the native advantages of transformers such as training efficiency and enhances their transfer learning capabilities through label-aware contrastive training.


\begin{table*}[tb]
\centering
\resizebox{1.0\textwidth}{!}{
\begin{tabular}{ccccccccccc}
\toprule
{ }                                                       & { }                                                          & { }                   & { \textbf{CIFAR-10}}   & { \textbf{CIFAR-100}}  & { \textbf{Cub-200-2011}}   & { \textbf{Oxford-Flowers}} & { \textbf{Oxford-Pets}} & { \textbf{iNat 2017}} & { \textbf{ImageNet-1k}} &  { \textbf{Places365}} \\ 
\multicolumn{1}{c}{{ \textbf{Model}}}    & \multicolumn{1}{c}{{ \textbf{Seen dataset}}} & { \textbf{FT method}} & { Acc@1} & { Acc@1} & { Acc@1} & { Acc@1}  & { Acc@1} & { Acc@1} & { Acc@1}  & { Acc@1} \\ \hline
\multicolumn{1}{c}{{ Data2vec}} & \multicolumn{1}{c}{{ ImageNet-21k}} & { CE}                 & { 98.25}      & { 89.21}      & { 85.16}      & { 91.57}          & { 94.52} & {71.05} & \textbf{84.20}  & {58.73}  \\ 
\hline
\multicolumn{1}{c}{{ ViT-B}}    & \multicolumn{1}{c}{{ ImageNet-1k}}  & { CE}                 & { 98.13}      & { 87.13}      & { N/A}        & { 89.49}          & { 93.81}  & {65.26} & {77.91}   & {54.06} \\ 
\multicolumn{1}{c}{{ \model{}-ViT-B}}    & \multicolumn{1}{c}{{ ImageNet-21k}} & { \model{}}           & { 99.07}      & { 91.01}      & { 85.69}      & \textbf{ 94.98}          & { 94.57}  & {70.38} & {82.99}  & {57.73}  \\ 

\multicolumn{1}{c}{{ ViT-L}}    & \multicolumn{1}{c}{{ ImageNet-1k}}  & { CE}                 & { 97.86}      & { 86.36}      & { N/A}        & { 89.66}          & { 93.64}   & {64.82} & {76.53} & {54.55}  \\ 
\hline
\multicolumn{1}{c}{{ SimMIM}}   & \multicolumn{1}{c}{{ ImageNet-1k}}  & { CE}                 & { 98.78}      & { 90.26}      & { 76.47}      & { 83.46}          & { 94.22}  & {70.28} & {83.00}  &  {57.54}\\

\multicolumn{1}{c}{{ \model{}-SimMIM}}   & \multicolumn{1}{c}{{ ImageNet-1k}}  & { \model{}}           & { 99.11}      & { 90.80}      & { 85.79}      & { 92.25}          & { 94.85} & {71.34} & {83.64}   & {58.47}  \\
\hline
\multicolumn{1}{c}{{ MAE}}      & \multicolumn{1}{c}{{ ImageNet-1k}}  & { CE}                 & { 98.28}      & { 87.67}      & { 78.46}      & { 91.67}          & { 94.05}  & {70.50} & {83.60}  &  {57.90} \\

\multicolumn{1}{c}{{ MAE}}      & \multicolumn{1}{c}{{ ImageNet-1k}}  & { SimCLR}             & { 97.53}      & { 76.01}      & { 57.91}      & { 89.22}          & { 91.15} & {65.62} & {81.92}   &  {55.48} \\
\multicolumn{1}{c}{{ MAE}}      & \multicolumn{1}{c}{{ ImageNet-1k}}  & { N-pair-loss}        & { 95.23}      & { 73.76}      & { 52.56}      & { 89.87}          & { 87.12} &  {62.36}  & {78.34} & {52.97}  \\

\multicolumn{1}{c}{{\model{}-MAE}}   & \multicolumn{1}{c}{{ ImageNet-1k}}  & { \model{}}           & \textbf{ 99.34}      & \textbf{ 91.27}      & \textbf{ 89.24}      & {93.34}          & \textbf{ 95.63}  &  \textbf{72.55} & { 84.12}  & \textbf{58.92} \\ 
\bottomrule
\end{tabular}
}
\vspace{-1mm}
\caption{\textbf{Image classification performance benchmarks over eight datasets.} CE refers to fine-tune with cross-entropy, while \model{} refers to fine-tune with our proposed label-aware contrastive fine-tuning framework. }
\label{tab:all_result}
\vspace{-2mm}
\end{table*}

\section{Experiments}
\label{sec:experiments}

\noindent\textbf{Experimental Setup}:
To evaluate the effectiveness of the \model{} framework, we conducted experiments using three state-of-the-art pretrained vision transformer models across eight diverse image classification datasets. The train/test splits for these datasets are consistent with prior work~\cite{RN83}. For the contrastive training stage, we initialize the encoder with pretrained weights obtained from either the ImageNet-1k or ImageNet-21k datasets. During training, we employ a batch size of 4096 for the contrastive training stage (Stage 1) and 128 for the task head fine-tuning stage (Stage 2). The number of epochs for these stages is set to 50 and 10, respectively. Both stages use an initial learning rate of \(1 \times 10^{-4}\). The temperature parameter \(\tau\) for the contrastive loss is set to 0.1. All the code used in our experiments can be found in \url{https://github.com/longkukuhi/LaCViT}.

\vspace{1mm} \noindent\textbf{Comparative Analysis}:
We benchmark the performance of \model{}-trained models against baseline models that solely utilize vanilla cross-entropy loss. The evaluation metrics include Top-1 accuracy across the selected datasets, as summarized in Table~\ref{tab:all_result}. Our results indicate that \model{}-trained models consistently outperform their baseline counterparts. For instance, \model{}-ViT-B achieves a Top-1 accuracy improvement of $3.88\%$ and $5.49\%$ on the CIFAR-100 and Oxford 102 Flowers datasets, respectively. \model{}-SimMIM shows a significant advantage over SimMIM, with an average improvement of $2.74\%$ over tested datasets. Data2vec performs worse than \model{}-MAE on smaller datasets such as CUB-200-2011 but shows marginal improvement on larger datasets, attributable to its larger pre-training dataset (ImageNet-21k). Moreover, \model{}-MAE emerges as the best-performing model on almost all datasets, with a performance boost of $10.78\%$ on the CUB-200-2011 dataset\footnote{An exception is the Oxford-Flowers dataset, where \model{}-ViT-B excels due to its ImageNet-21k pre-training.}.

\vspace{1mm} \noindent\textbf{Discussion}:
The observed performance gains substantiate the effectiveness of our label-aware contrastive training approach in \model{}. This is particularly prominent in smaller datasets but is also evident in larger datasets, such as iNat 2017 and ImageNet-1k. \model{}-MAE achieves better performance with less pre-training data, which demonstrates that extensive pre-training on larger datasets does not necessarily translate to improved transferability on smaller datasets. With \model{}, comparable or even superior performance can be attained.

\section{Analysis}

\noindent\textbf{Ablation Study on Alternative Contrastive Loss}: 
We evaluate the performance of \(\text{\model{}}\) against other notable unsupervised contrastive learning methods, namely SimCLR~\cite{RN89} and N-pair-loss~\cite{npairloss}, to understand the unique advantages of our label-aware approach. We use the MAE base model for these comparisons. The results are summarized in the lower section of Table~\ref{tab:all_result}. MAE fine-tuned with SimCLR significantly underperforms compared to \(\text{\model{}}\)-MAE, particularly on the CUB-200-2011 and CIFAR-100 datasets, registering a performance decrease up to $31.33\%$. When fine-tuned with N-pair-loss, MAE exhibits a 2--5\% decline in accuracy compared to its SimCLR counterpart, with the exception of a marginal accuracy gain of 0.65\% on the Oxford 102 Flower dataset. These findings suggest that unsupervised contrastive learning methods may not sufficiently capture class-specific features, thus affecting performance adversely. Cross-entropy fine-tuned MAE consistently outperforms both SimCLR and N-pair-loss fine-tuned versions, emphasizing the need for using label information in contrastive learning like \(\text{\model{}}\).

\begin{figure}[h]
    \begin{subfigure}{0.45\linewidth}
        \includegraphics[width=1\linewidth]{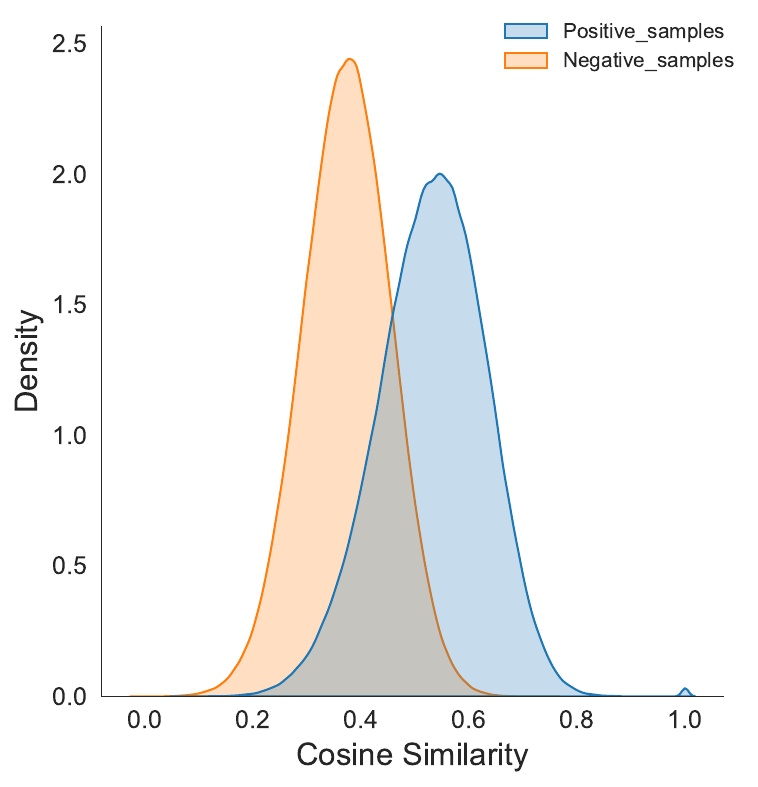}
        \subcaption[]{MAE}
        \label{fig:MAE_cosine}
    \end{subfigure}
\hfill 
    \begin{subfigure}{0.45\linewidth}
        \includegraphics[width=1\linewidth]{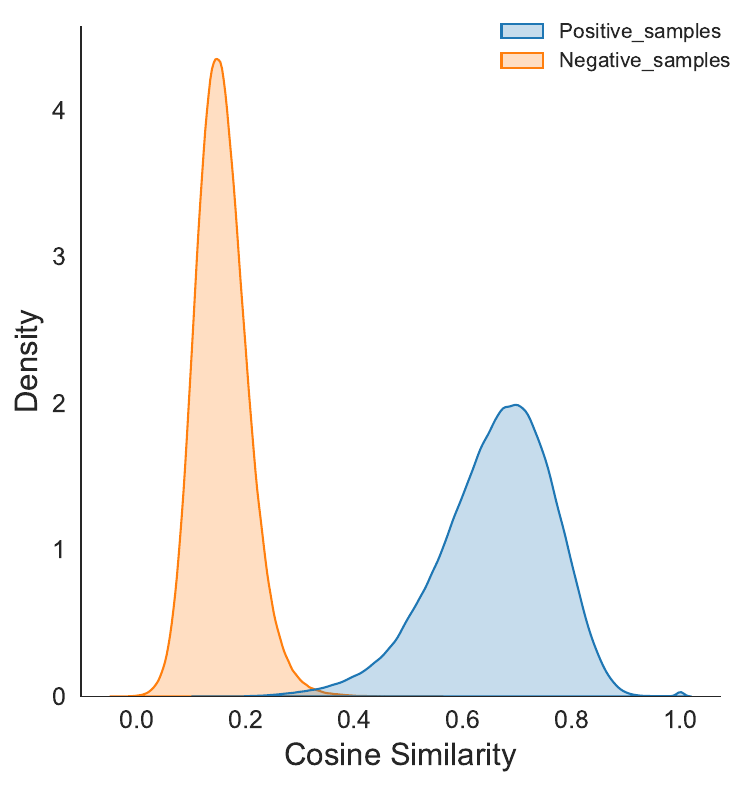}
        \subcaption[]{\model{}-MAE}
        \label{fig:conmae_cosine}
    \end{subfigure}
\caption{\textbf{Plot of cosine similarity distribution across two random classes from CIFAR-10.} Blue and orange mean positive and negative similarities, respectively.}
\label{fig:cosine}
\vspace{-4mm}
\end{figure}

\begin{figure}[h]
\vspace{-2mm}
    \begin{subfigure}{0.48\linewidth}
        \includegraphics[width=1\linewidth]{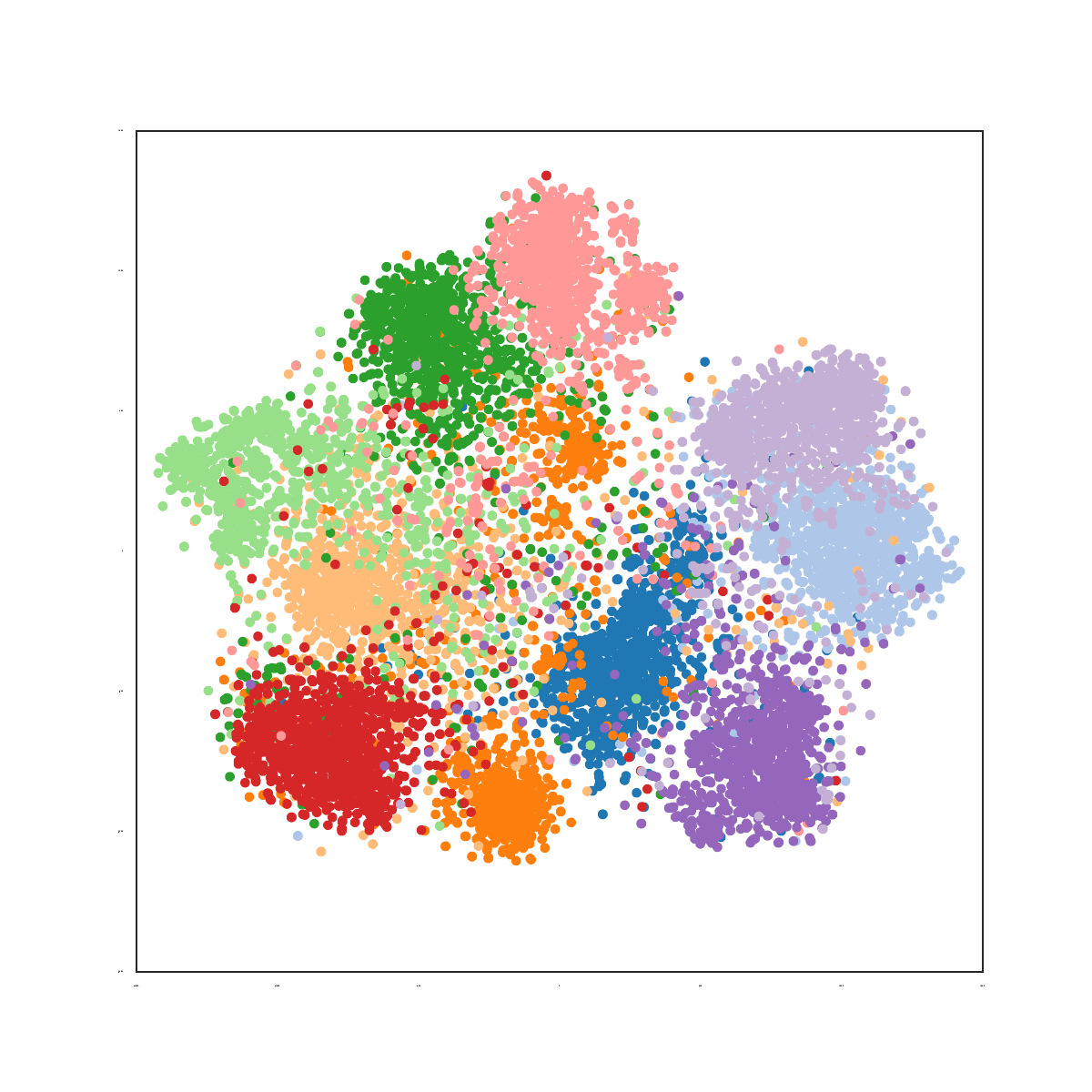}
        \subcaption[]{MAE}
        \label{fig:MAE_tsne}
    \end{subfigure}
\hfill 
    \begin{subfigure}{0.48\linewidth}
        \includegraphics[width=1\linewidth]{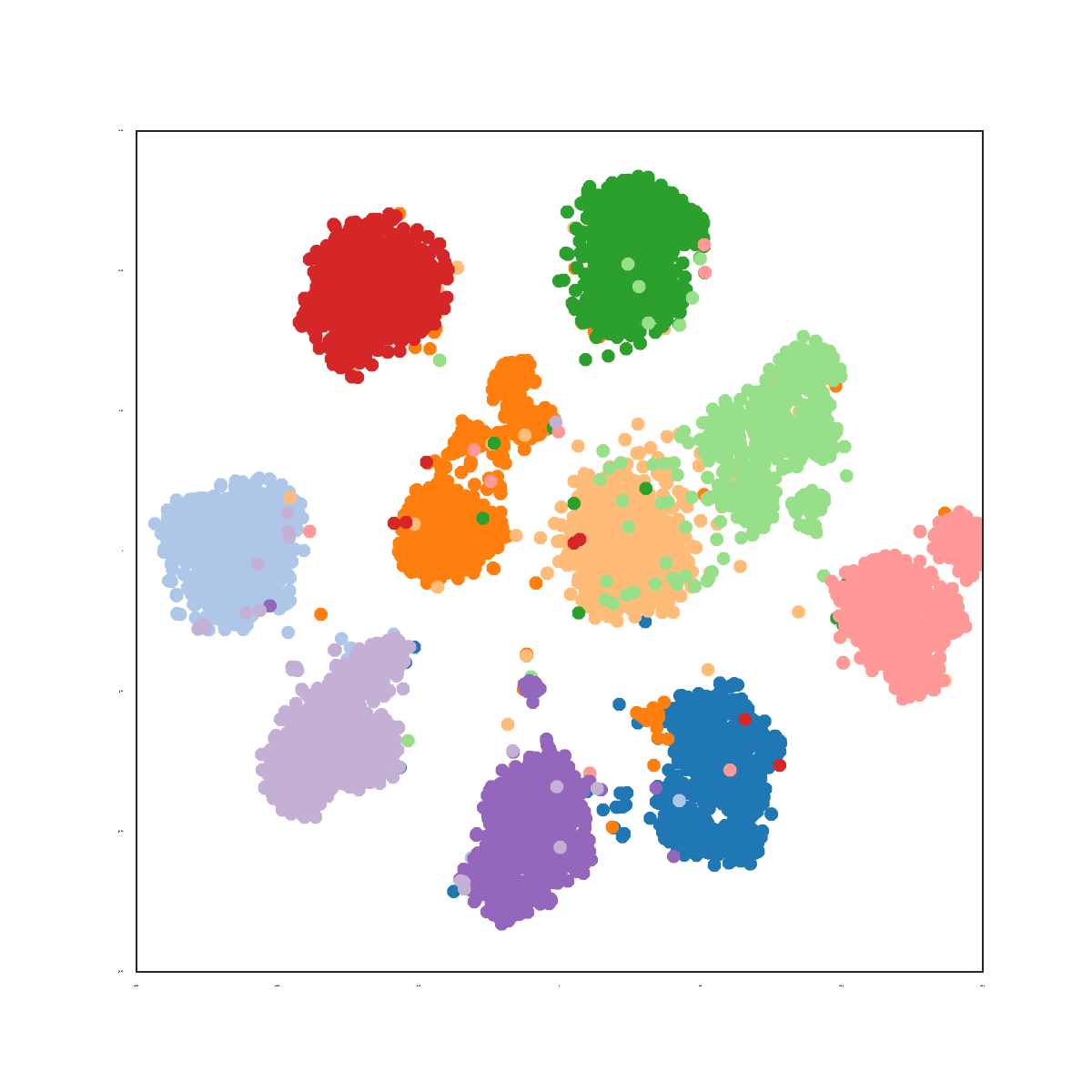}
        \subcaption[]{\model{}-MAE}
        \label{fig:conmae_tsne}
    \end{subfigure}
\caption{\textbf{Embedding Space Visualization for MAE vs. \model{}-MAE}. Displayed over ten CIFAR-10 classes using t-SNE. Each dot represents a sample, with distinct colors indicating different label classes.}
\label{fig:tsne}
\vspace{-4mm}
\end{figure}

\vspace{1mm} \noindent \textbf{Embedding Quality Analysis}: 
We further investigate the geometric properties of the learned embedding spaces to understand the impact of label-aware contrastive training.
\vspace{-2mm}
\begin{itemize}
\item\textbf{Cosine Similarity}: Figure~\ref{fig:cosine} illustrates the cosine similarity distribution between \(\text{\model{}}\)-MAE and MAE. The figure shows that \(\text{\model{}}\)-MAE offers better inter-class separation. 
\vspace{-2mm}\item \textbf{t-SNE Visualization}: Figure~\ref{fig:tsne} presents t-SNE visualizations of the embeddings for both MAE and \(\text{\model{}}\)-MAE. The clusters in \(\text{\model{}}\)-MAE are tighter and better separated, thus underscoring the discriminative power of label-aware contrastive training.
\end{itemize}

\vspace{1mm} \looseness -1 \noindent \textbf{Summary}: Our analysis confirms that label-aware contrastive training with \(\text{\model{}}\) enhances the geometric properties of the embedding spaces, particularly in terms of inter-class separation. These improvements substantiate the superior transfer learning capabilities of vision transformers trained using \(\text{\model{}}\).

\vspace{-2mm}
\section{Conclusions}
\looseness -1 We present \model{}, a label-aware contrastive fine-tuning framework that significantly increases the Top-1 accuracy of vision transformers across various benchmarks. It outperforms state-of-the-art models like MAE by up to 10.78\% and is applicable to other transformers such as ViT and SimMIM. Through rigorous analysis, including using cosine similarity metrics and t-SNE visualizations, we demonstrate that \model{} effectively reshapes the geometric properties of the embedding space, contributing to its effectiveness in image classification tasks. In summary, \model{} offers a comprehensive and versatile approach that serves to substantially elevate the utility of transformers in image classification. Our exhaustive empirical evaluations not only validate the effectiveness of \model{} but also suggest that it offers an effective alternative to cross-entropy for fine-tuning pretrained models for image classification tasks.

\bibliographystyle{IEEEbib}
\bibliography{strings,refs}

\end{document}